\documentclass{SCIS2025}

\begin{document}
\ArticleType{REVIEW}
\Year{2025}
\Month{}
\Vol{}
\No{}
\DOI{}
\ArtNo{}
\ReceiveDate{}
\ReviseDate{}
\AcceptDate{}
\OnlineDate{}
\AuthorMark{}
\AuthorCitation{}

\title{Satellite edge artificial intelligence with large models: Architectures and technologies}{Y. Shi, J. Zhu, C. Jiang, L. Kuang, and K. B. Letaief, Satellite edge artificial intelligence with large models: Architectures and technologies}

\author[1]{Yuanming Shi}{}
\author[1]{Jingyang Zhu}{{zhujy2@shanghaitech.edu.cn}}
\author[2]{Chunxiao Jiang}{}
\author[2]{Linling Kuang}{}
\author[3]{Khaled B. Letaief}{}


\address[1]{School of Information Science and Technology, ShanghaiTech University, Shanghai {\rm 201210}, China}
\address[2]{Beijing National Research Center for Information Science and Technology, Tsinghua University, Beijing {\rm 100084}, China}
\address[3]{Department of Electronic and Computer Engineering, Hong Kong University of Science and Technology, Hong Kong}

\abstract{Driven by the growing demand for intelligent remote sensing applications, large artificial intelligence (AI) models pre-trained on large-scale unlabeled datasets and fine-tuned for downstream tasks have significantly improved learning performance for various downstream tasks due to their generalization capabilities. However, many specific downstream tasks, such as extreme weather nowcasting (e.g., downburst and tornado), disaster monitoring, and battlefield surveillance, require real-time data processing. 
Traditional methods via transferring raw data to ground stations for processing often cause significant issues in terms of latency and trustworthiness.
To address these challenges, satellite edge AI provides a paradigm shift from ground-based to on-board data processing by leveraging the integrated communication-and-computation capabilities in space computing power networks (Space-CPN), thereby enhancing the timeliness, effectiveness, and trustworthiness for remote sensing downstream tasks. Moreover, satellite edge large AI model (LAM) involves both the training (i.e., fine-tuning) and inference phases, where a key challenge lies in developing computation task decomposition principles to support scalable LAM deployment in resource-constrained space networks with time-varying topologies.
In this article, we first propose a satellite federated fine-tuning architecture to split and deploy the modules of LAM over space and ground networks for efficient LAM fine-tuning. 
We then introduce a microservice-empowered satellite edge LAM inference architecture that virtualizes LAM components into lightweight microservices tailored for multi-task multimodal inference. 
Finally, we discuss the future directions for enhancing the efficiency and scalability of satellite edge LAM, including task-oriented communication, brain-inspired computing, and satellite edge AI network optimization.}

\keywords{Large AI models (LAMs), satellite edge LAM, space-computing power networks (Space-CPN), satellite federated fine-tuning, and microservice-empowered satellite edge LAM inference.}

\maketitle

\section{Introduction}
Remote sensing (RS) satellite Earth observation systems enable global coverage, all-weather, and full-spectrum detection capabilities, playing a crucial role in global economic development, resource management, and environmental protection. 
Multi-resolution, multi-sensor, and multi-temporal unlabeled RS data can provide different temporal-spatial-spectral descriptive information of objects, empowering various downstream tasks such as environmental monitoring, extreme weather forecasting, disaster monitoring, and military surveillance \cite{sun2023single}. Artificial intelligence (AI) methods, represented by deep learning, have significantly improved the efficiency of RS data interpretation. 
Currently, an important development direction in the field of intelligent RS data interpretation is to leverage large-scale multimodal data, enabling AI models to have multi-task generalization capabilities while achieving efficient and high-precision inference. The approach of pre-training large AI models (LAMs) followed by fine-tuning for downstream tasks is a proven effective solution. RS large models, which are pre-trained on large-scale unlabeled datasets (e.g., fMoW \cite{christie2018functional} and BigEarthNet \cite{sumbul2019bigearthnet}) using self-supervised learning techniques, have achieved significant performance improvement. Existing LAMs have adopted various neural network architectures, including diffusion models (e.g., DiffusionSat \cite{khanna2024diffusionsat}) and Transformer-based models (e.g., SatMAE \cite{cong2022satmae} and SpectralGPT \cite{hong2024spectralgpt}).

\begin{figure}[!]
	\centering
	\includegraphics[width=1\linewidth]{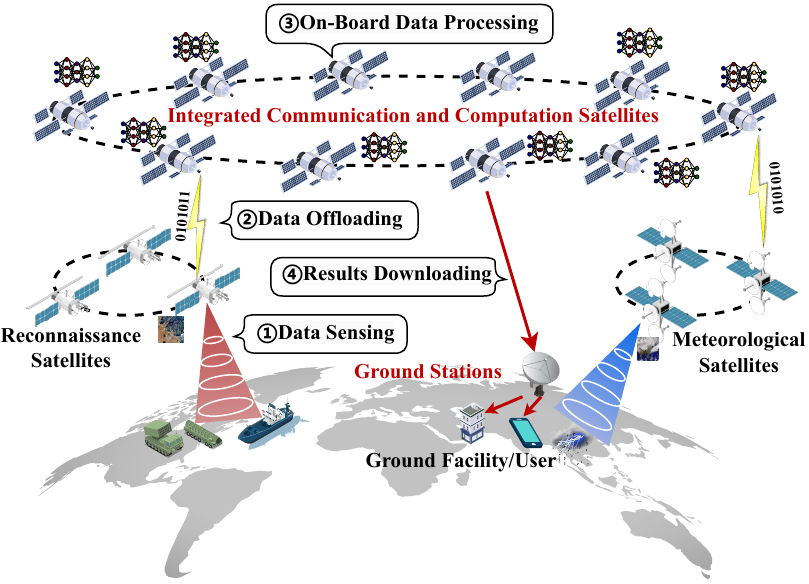}
	\caption[]{The procedures of satellite edge AI including data sensing, data offloading, on-board data processing, and results downloading.}
	\label{fig: edge AI}
\end{figure}

Many specific RS downstream tasks have high requirements for data timeliness.
For instance, in the field of extreme weather forecasting, severe convective weather events such as downbursts, tornadoes, and hailstorms are characterized by sudden onset, strong locality, short lifespan, and significant impact, making them a persistent challenge in both weather forecasting (i.e., on the scale of hours) and nowcasting tasks (i.e., on the scale of minutes) \cite{lebedev2019precipitation}.
Directly downloading such massive RS data for ground processing would inevitably lead to severe latency issues and pose risks of privacy breaches \cite{coffer2020balancing}.
For instance, in 2022 alone, Landsat-8 from NASA's Landsat program generated 73.8 TB of publicly available RS imagery over regions in China.
Meanwhile, the transmission rates of common feeder links typically range from several hundred Mbps to a few Gbps \cite{horst2023tbit}. 
Therefore, developing more efficient architectures for scalable and low-latency intelligent data processing becomes critical.
To address this substantial demand for computing power, the space computing power networks (Space-CPN) have emerged as a promising solution to provide global connectivity and support ubiquitous intelligent services, encompassing aircraft and various satellites in space. 
{The on-board processing capabilities of satellites have significantly improved, particularly with respect to graphics processing units (GPUs) designed for neural network training. 
The GPU for Space program, supported by the European Space Agency and led by the Barcelona Supercomputing Center, successfully validated the feasibility, low power consumption, and high performance of integrating GPUs on satellites in 2019 \cite{kosmidis2020gpu4s}. 
Besides, European Space Agency’s $\Phi$-Sat-1 mission marked the first successful use of deep neural networks to assist Earth observation tasks \cite{giuffrida2022sat1}.
}
In this context, complex RS downstream services at the satellite edge can be enhanced by leveraging LAMs, which benefit from the continuously advancing computational and communication capabilities of nodes within Space-CPN.
The overall architecture of satellite edge AI is illustrated in Fig. \ref{fig: edge AI}. Satellite edge AI with LAMs primarily involves two major types of computation tasks: training and inference.

Training tasks in satellite edge LAM refers to fine-tune the LAMs using the satellite sensing data to enhance the timeliness and trustworthiness for the downstream tasks. This process faces the challenge of limited computing power at the satellite edge, making it difficult to meet the high computation demands of large model fine-tuning tasks even with the collaboration among multiple satellites. Therefore, exploring the computation task decomposition approaches of satellite fine-tuning tasks for LAMs and developing joint communication and computation orchestration methods are key to addressing the issue of insufficient single-point computing power and supporting on-board fine-tuning \cite{lin2024split,you2025ai}.
In this article, we shall propose a satellite federated fine-tuning architecture, establishing an intelligent coordination mechanism between Space-CPN elements and ground facilities to enhance the timeliness of satellite fine-tuning tasks and improve the efficiency of network resource utilization. 
Specifically, the LAM modules are partitioned, with the parameter-heavy components deployed on the ground cloud server, while the satellites utilize their computing payloads to fine-tune the classification layers of the LAM. This requires designing new inter-satellite scheduling schemes and satellite-ground coordinated transmission technologies to support the satellite federated fine-tuning.

Satellite edge LAM inference refers to provide on-board interpretation empowered by LAMs for various RS downstream applications requested from ground users, thereby improving the overall inference timeliness and accuracy \cite{yu2024microservice}.
These multimodal inference tasks involve the sequential processing of requested data through the main modules of LAMs, including modality encoders, input projectors, a backbone calculator, an output projector, and a modality decoder. 
Many downstream inference tasks require the use of the same multiple modality encoders and input projectors, resulting in substantial computation demands and increased latency. This however leads to significant computation redundancy, where identical modality encoders and mappers are repeatedly scheduled across different inference tasks. Furthermore, in traditional multimodal multitask inference, where different modules are consolidated into a single multimodal network for a specific inference task, there is often redundant deployment of common encoding and projection modules for different tasks \cite{Al2023AI}.
To address these issues, we shall propose a microservice-empowered satellite edge LAM inference architecture for multi-task multi-modal inference. In this architecture, the functional modules (e.g., modality encoders, input projectors) of the LAM are virtualized into a series of independent microservices, presenting the advantages of low coupling and high cohesion. 
However, due to the time-varying property of satellite network topology and the heterogeneity of computation resources, efficiently deploying microservice modules and orchestrating them sequentially at the satellite edge remain critical challenges. This paper discusses intuitive methods for microservice deployment and orchestration at the satellite edge to address these challenges.

\subsection{Contributions and Organizations}
The main contributions of this article are three-fold. First, we propose a satellite federated fine-tuning architecture tailored for satellite edge training by partitioning and distributing the modules of LAMs and developing corresponding inter-satellite link scheduling and satellite-ground coordinated transmission schemes to support this architecture in Section \ref{sec: training}. 
To support inference service for various multimodal downstream tasks, we introduce a microservice-empowered satellite edge LAM inference architecture to address computation redundancy problem originated from multi-task multimodal LAM inference, followed by providing solutions for the microservice deployment and orchestration problems over Space-CPN in Section \ref{sec: inference}. 
Finally, we outline our future work from three perspectives: task-oriented communications, neuromorphic computing, and generative AI for optimizing satellite edge AI networks in Section \ref{sec: future}.

\section{Satellite Federated Fine-Tuning for Large AI Model Training}\label{sec: training}
In this section, we present the satellite federated fine-tuning architecture, followed by introducing the inter-satellite link scheduling schemes and the satellite-ground coordinated transmission techniques.
\subsection{Satellite Federated Fine-Tuning Architecture}
With the deployment of numerous RS satellite projects such as the Landsat series, Gaofen series, and Sentinel-2, satellites equipped with various types of sensors can capture massive multimodal data, including multispectral, hyperspectral, infrared, and synthetic aperture radar images. These RS data can support a wide range of Earth observation applications, including land use classification, weather forecasting, and disaster monitoring. The success of RS large models are demonstrated by their significant enhancement of traditional data-driven models' feature representation capabilities and their improved generalization performance on downstream tasks. We shall utilize the state-of-the-art RS large model, SpectralGPT \cite{hong2024spectralgpt}. SpectralGPT adopts the Vision Transformer (ViT) \cite{dosovitskiy2021an} as its network backbone and is pre-trained on a bunch of large-scale multispectral dataset comprising millions of images with varying sizes, time series information, and earth regions. A ViT consists of an embedding layer for image patching and position embedding, several Transformer encoders, and a multi-layer perceptron (MLP) head for classification. Typically, ViT has three versions with different parameter sizes: ViT-Base with 86 million parameters, ViT-Large with 307 million parameters, and ViT-Huge with 632 million parameters. Specifically, masked autoencoder (MAE) \cite{he2022masked}, an emerging baseline method, are used in the pre-training phase of SpectralGPT, where unmasked patches or pixels are employed to reconstruct the masked ones in a self-supervised manner. As a result, in the pre-training phase, an encoder for learning visual representations and a decoder for image reconstruction can be obtained, where the decoder replaces the MLP head during pretraining, and the MLP head is used for fine-tuning in downstream tasks.
The RS large models can be then applied to other downstream tasks by fine-tuning them on specific dataset, which aims at enhancing the learning performance by updating the parameters of the LAM, instead of training from the very beginning \cite{dong2024upentu}. 

A traditional way to pre-train and fine-tune LAMs using massive space data is downloading them to the ground facilities for centralized training. However, downloading raw satellite data violates some data privacy protocols, as well as suffering from huge communication overhead. For instance, the European Space Agency's Sentinel-2, as part of its Earth observation satellite constellation, is expected to generate 1.6 TB of high-resolution optical imagery data within each observation cycle across its various orbits. To this end, we shall propose a satellite federated fine-tuning scheme to update local models on-board without transmitting raw data to ground stations (GS). However, fine-tuning the full LAM (e.g., SpectralGPT with ViT-Huge backbone has 632 million parameters) on-board results in extremely high computation and communication complexity and storage pressure for resource-constraint satellites. Parameter-efficient fine-tuning (PEFT) methods such as prompt tuning, prefix tuning, adapter tuning, and low-rank adapters, have been proposed to reduce the number of fine-tuning parameters for pre-trained LAM while achieving comparable learning performance with fine-tuning full parameter LAM \cite{jiao2023brain}. Instead, we propose to adopt parameter-efficient head tuning, where the parameters of the encoder in the foundation model are kept frozen, and only the head layer is fine-tuned for different downstream tasks. For example, in single- and multi-label classification tasks, the original MLP can be fine-tuned, while for tasks such as semantic segmentation or change detection, UperNet layers can be added after the MLP layer for head tuning \cite{hong2024spectralgpt}. Given the fact that the processing capabilities of the satellites are limited, whereas ground cloud resources are abundant, we propose a unique model deployment scheme tailored for satellite federated head fine-tuning, where the embedding layer and head layer of the foundation model SpectralGPT \cite{hong2024spectralgpt} are deployed on the LEO satellites, while the encoder, which contains the majority of the model's parameters, is deployed on the ground cloud server. To elaborate, the embedding and head layers of SpectralGPT  have around 50,000 and 62,000 parameters, respectively, accounting for only 0.13\% of the total parameters in ViT-Base. In this setup, the satellite federated fine-tuning task aims at training the local head layer on each LEO satellite using their local data and obtaining a global head after model aggregation.
\begin{figure*}[t]
	\centering
	\includegraphics[width=1\linewidth]{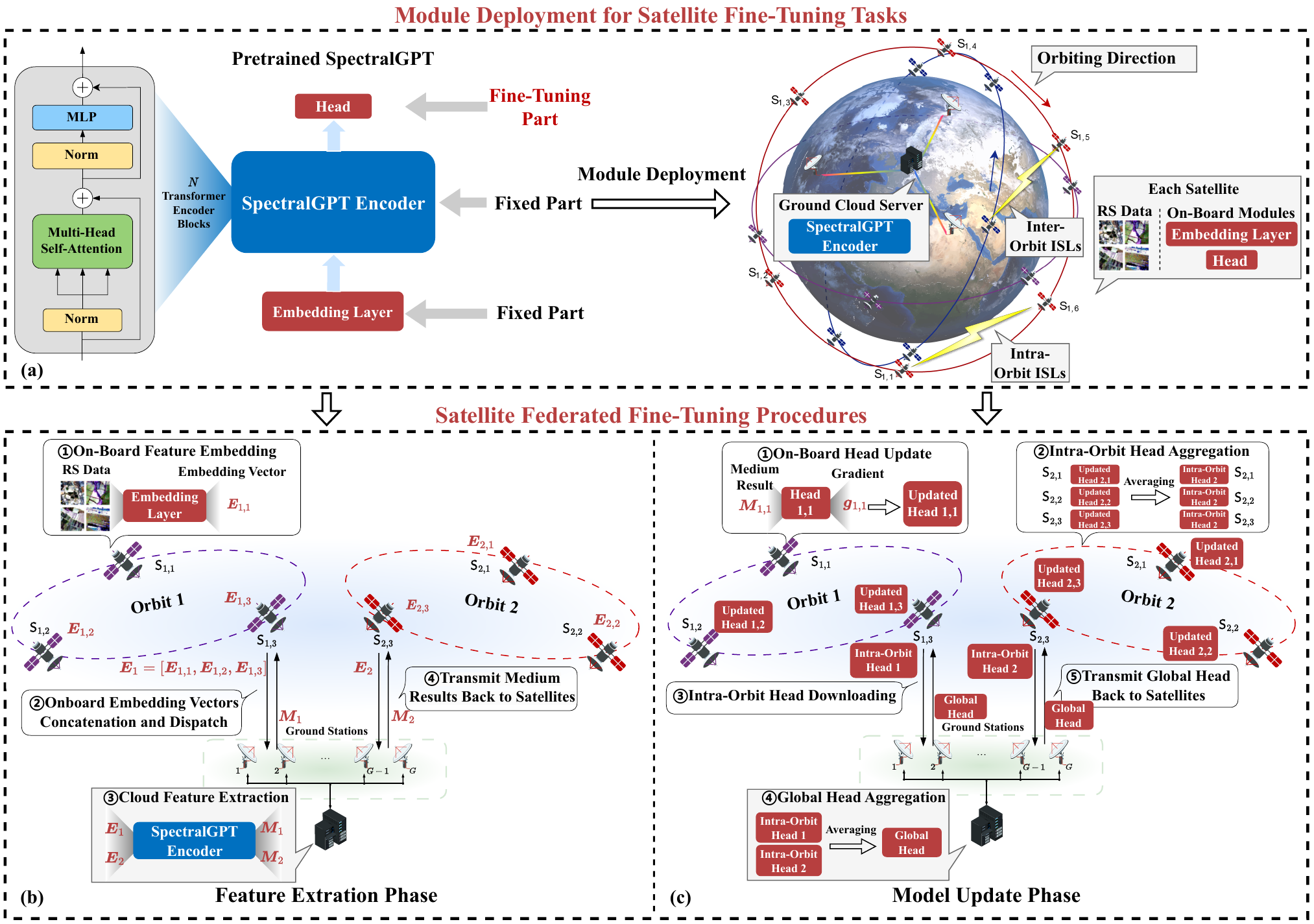}
	\caption[]{The satellite federated fine-tuning architecture, including LAM module deployment and operation procedures.}
	\label{fig: finetuning}
\end{figure*}

The detailed communication and computation process of the proposed satellite federated fine-tuning architecture is illustrated in Fig. \ref{fig: finetuning}, which consists of two phase, i.e., feature extraction in Fig. \ref{fig: finetuning}(b) and model update in Fig. \ref{fig: finetuning}(c). In the feature extraction phase, each LEO satellite first computes embedding vectors by transforming local data into patches, which are then input into the embedding layer. Subsequently, one LEO satellite within each orbit gathers the embedding vectors from all satellites via intra-orbit inter-satellite links (ISL) and sends the concatenated intra-orbit embedding vectors to the GSs via satellite-ground links (SGL). GSs relay the concatenated embedding vectors to the ground cloud server via ground dedicated links. The ground cloud server then extracts feature vectors from the embedding vectors via the deployed encoder and sends them back. Once the LEO satellites receive the feature vectors, the model update phase begins. The LEO satellites first train their local heads and then perform a intra-orbit model aggregation via ISLs to accelerate the model fine-tuning. After several intra-orbit communication rounds, each satellite in one orbit obtains the same intra-orbit head and delivers the updated head to the GSs for global aggregation via SGL. Finally, a global head is broadcast to all satellites for training in the next communication round. This process repeats until the fine-tuning procedure converges. 

\subsection{Inter-Satellite Link Scheduling and Satellite-Ground Coordinated Transmission}
Specifically, in the satellite federated fine-tuning architecture, although it is unnecessary to transmit terabytes of data to the ground cloud server, several challenges remain to be tackled. First, while the embedding layer placed on the LEO satellite has a relatively small number of parameters, the output embedding vectors can be substantial due to the large number of high-resolution images resized and input into the embedding layer of SpectralGPT. For example, using the EuroSAT dataset with a batch size of 512, the output embedding vector dimension is 786,432, and when stored with 32-bit float precision, each batch consumes approximately 1.5 GB of memory. Given that the Ka/Ku-band feeder links between LEO satellites and GSs generally have capacities ranging from several hundred Mbps to a few Gbps, and that satellite-ground link durations typically last only a few minutes, downloading all embedding vectors from satellites in an orbit to the ground still presents significant challenges \cite{horst2023tbit}. 
Second, with the development of laser ISL technology \cite{wang2024free}, LEO satellites can transfer data between one another at speeds of several to tens of Gbps via laser ISLs. 
Effectively leveraging inter-satellite communication during the feature extraction phase for transmitting embedding vectors, and utilizing intra-orbit aggregation during the model update phase to accelerate the convergence of fine-tuning tasks, is crucial. 
In summary, it is essential to propose new task-oriented inter-satellite and satellite-ground transmission approaches to support satellite federated fine-tuning tasks.

\subsubsection{Inter-Satellite Link Scheduling Schemes}
During the satellite federated fine-tuning process, massive data is exchanged through intra-orbit ISLs, including embedding vectors and local heads. 
To improve the transmission efficiency and accelerate model fine-tuning, we propose a ring all-reduce based algorithm to support both intra-orbit embedding gathering and model aggregation, which leverages the stable ring topology of LEO satellites in one orbit. 
First, the on-board data is divided into several blocks. 
Next, each satellite transmits a certain block to its neighbor in parallel. 
At the end of transmission, each satellite has the complete data of two certain blocks. 
To proceed, each satellite transmit the local complete data block to its neighborhood until each satellite possesses information from all satellites in the orbit. 
With this parallel transmission scheme, the transmission time is no longer related to the number of LEO satellites in one orbit, but only to the size of the data. It is worth noting that the rate of ISLs may vary between different satellites or change over time in real systems, and the overall performance is constrained by the slowest link. Additionally, ISLs between adjacent satellites within the same orbit may also be affected by alignment losses, leading to potential transmission outages \cite{nie2021channel}.

The global model aggregation with a ground cloud server as the central coordinator is limited by the low data rates of SGLs, as well as single-point failure issues, meaning that if the ground cloud server fails, the training process cannot continue. 
To solve this problem, the decentralized head fine-tuning can also be considered with complete on-board model aggregation via intra-orbit and inter-orbit ISLs \cite{wang2024free}. Existing decentralized designs on the ground cannot be directly utilized in LEO satellite networks due to the special topology of satellite networks. Specifically, high-speed and stable intra-orbit ISLs with the ring topology guarantee fast intra-orbit transmission \cite{wu2024scalable}. Furthermore, inter-orbit ISLs are time-varying and unreliable. For instance, in several constellations such as Kepler, OneWeb, and Starlink, the inter-orbit ISL rates primarily range from 1 to 4 Gbps \cite{leyva2022ngso}, which is significantly lower than that of intra-orbit ISLs. 
{ISLs established between satellites orbiting in opposite directions are referred to as cross-seam inter-orbit ISLs \cite{leyva2021inter}, which suffer from Doppler shift (reaching up to 2.4 GHz) and short duration due to the rapid movement of the satellites.
This results in an increased bit error rate during inter-orbit data transmission. Therefore, it is essential to strategically schedule the use of inter-orbit links to maintain the overall efficiency and reliability of the system.
} 
To accelerate inter-orbit transmission process, a set of paths is selected to transmit the model in parallel by setting the weight of each ISL as the reciprocal of the link capacity. Subsequently, we propose to utilize the Floyd-Warshall algorithm to find the shortest path between each pair of satellites \cite{gallo1988shortest}. The shortest path between the first orbit and the last orbit is selected and any path that overlaps the shortest path is removed to avoid congestion. Repeat this process and selected paths are used to transmit data in parallel. The tailored inter-orbit communication scheme solves the single-point failure problem and fully utilizes satellite transmission resources to transmit models in parallel to minimizes latency.

\begin{figure}[t]
	\centering
	\includegraphics[width=1\linewidth]{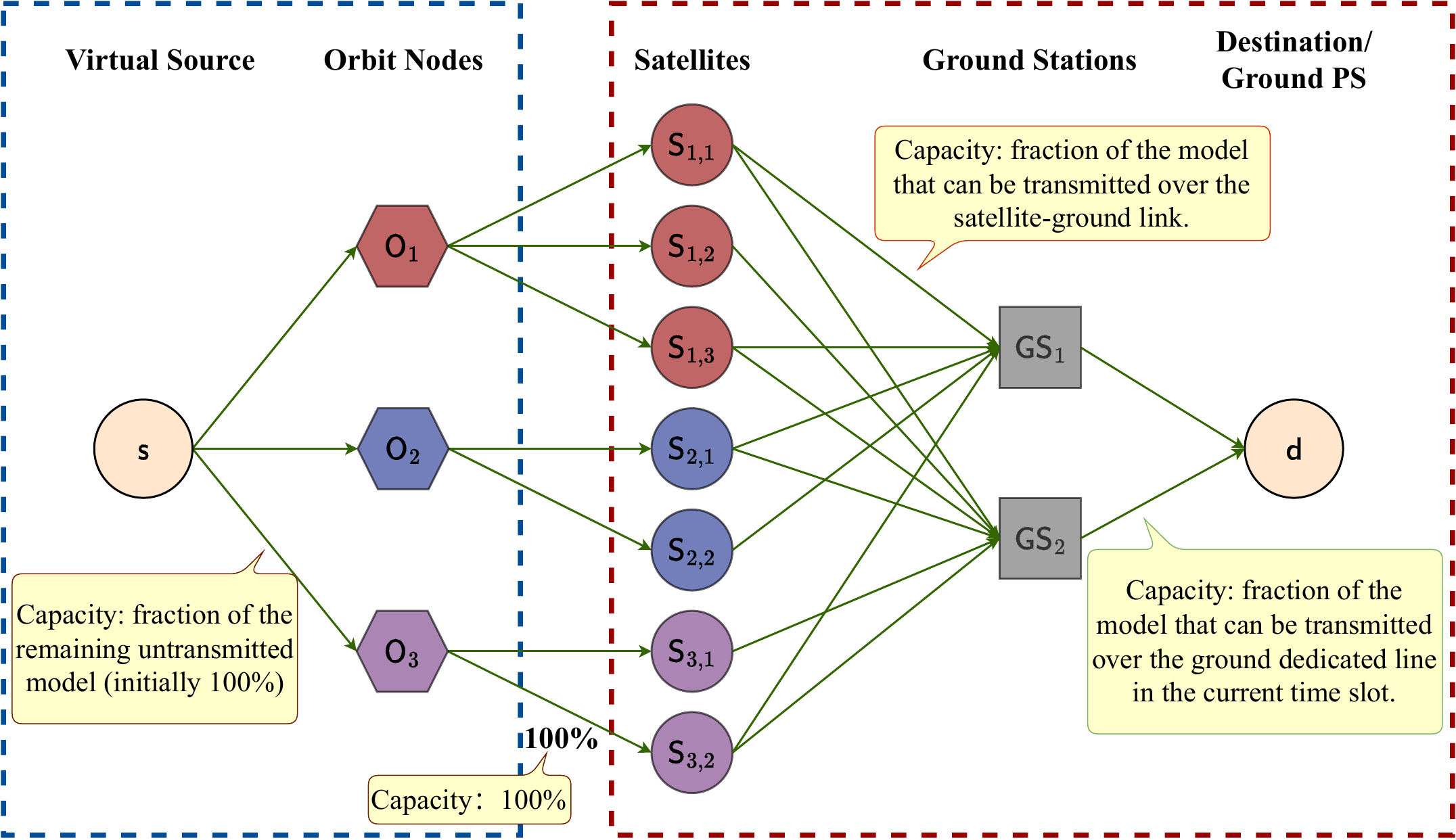}
	\caption[]{The formulated network flow problem for satellite-ground coordinated transmission.}
	\label{fig: networkflow}
\end{figure}

\subsubsection{Satellite-Ground Coordinated Transmission Schemes}
Through multi-beam and multiple access technologies, a single LEO satellite can simultaneously connect to multiple GSs, and a GS can establish SGLs with multiple satellites within its line of sight. Given that all satellites within the same orbit have an identical intra-orbit model upon completing intra-orbit aggregation, the model from each orbit can be transmitted collaboratively by several LEO satellites to multiple GSs, significantly reducing the data load required to be transmitted over each SGL. Specifically, the satellite-ground communication topology in each communication window can be modeled as a static flow network, which contains edges, satellite vertices, GS vertices, a sink vertex and a virtual source vertex. For an edge between the source vertex and a satellite vertex, its weight is determined by the fraction of untransmitted data in the corresponding orbit during the current communication window. For an edge between a satellite vertex and a GS vertex, its weight is determined by the maximum proportion of data that can be transferred via the corresponding link during the current communication window. For an edge between a GS vertex and the sink vertex, its capacity is determined by the maximum proportion of data that can be transferred via the dedicated link during the current communication window. This problem of maximizing the overall data rate during satellite-ground data transmission can be formulated as a classical network flow problem and solved by adopting the well-known Ford-Fulkerson algorithm \cite{shi2024satellite}. {The network flow model simulates the flow of water through pipes to identify the maximum data transmission path in a capacity-constrained network, ensuring that data is transmitted from satellites to GSs in the most efficient manner.} The network flow problem is demonstrated in Fig. \ref{fig: networkflow}.

Thanks to the rapid development and deployment of LEO satellite array antennas, SGL is increasingly adopting multiple-input-multiple-output (MIMO) communication technology \cite{heo2023mimo}. In addition, over-the-air computation (AirComp) emerges as a promising approach to achieve a more efficient aggregation process. AirComp is designed to compute a class of nomographic functions over wireless channels by leveraging the waveform superposition property of multiple access channels \cite{wang2024over}. With this technique, signals from each device are transmitted simultaneously, allowing the summation operation to be performed directly over the air. A potential application of AirComp could involve coordinated downlink transmission between LEO satellite swarm and GSs, which offers notable spectral efficiency improvements due to both the increased spatial separation of antennas and the inherent benefits of AirComp. However, the high mobility of LEO satellites presents new challenges for AirComp-based MIMO SGL systems, particularly in acquiring accurate channel state information (CSI) within the limited coherence time. To mitigate this, the predictable movement patterns of LEO satellites can be used to approximate CSI by considering the positional relationship between transmit and receive antennas, along with estimated atmospheric influences. This approach enables the integration of communication and computation, ultimately enhancing spectral efficiency and reducing aggregation latency for SGL systems. Nevertheless, affected by various factors such as strong interference, unstable weather conditions, and significant atmospheric attenuation, the communication link between the ground station and LEO satellites typically experiences a low signal-to-noise ratio (SNR). This inevitably leads to severe distortion if analog modulation-based AirComp is used to aggregate data at satellites. To address this issue, lattice quantization provides an effective solution for enabling coded AirComp, offering strong noise resistance while maintaining compatibility with existing digital systems \cite{azimi2024compute}. Moreover, the lattice coding approach for AirComp eliminates the need for CSI at the transmitter, which makes it particularly suitable for SGL systems. {To conclude, the lattice coding approach for AirComp integrates the benefits of both digital and analog transmission, as discussed in \cite{yao2024wireless}. This method enhances bandwidth efficiency, improves system robustness against channel conditions, and reduces latency.
}

\section{Microservice-Empowered Satellite Edge Inference for Large AI Model} \label{sec: inference}
In this section, we present the microservice-empowered satellite edge inference architecture for multi-task multimodal LAM inference, followed by developing microservice deployment and orchestration schemes over Space-CPN.
\subsection{Microservice-Empowered Satellite Edge LAM Inference Architecture}
Mainstream multimodal LAMs comprise modality encoders, input projectors, a backbone calculator, an output projector, and a modality decoder, facilitating various inference tasks through the sequential processing of requested data (e.g., text and image modal data) via these modules. For example, applications such as generating videos from text and images, removing clutter from images based on textual prompts, or intelligent interpretation of RS data all fall under the category of multimodal LAM inference tasks. Multi-task inference in multimodal LAMs necessitates the activation of multiple modality encoders and input projectors, resulting in substantial computation demands and increased latency. Existing approaches mitigate computation latency by minimizing the calculation overhead of the overall compressed model, which ignores the heightened response latency caused by the repeated scheduling of identical modality encoders and mappers across different inference tasks \cite{xu2023joint}. Besides, traditional multimodal multitask inference, which consolidates different modules into a single multimodal network for a specific inference task, there is often redundant deployment of common encoding and projection modules across various networks.
\begin{figure*}[t]
	\centering
	\includegraphics[width=1\linewidth]{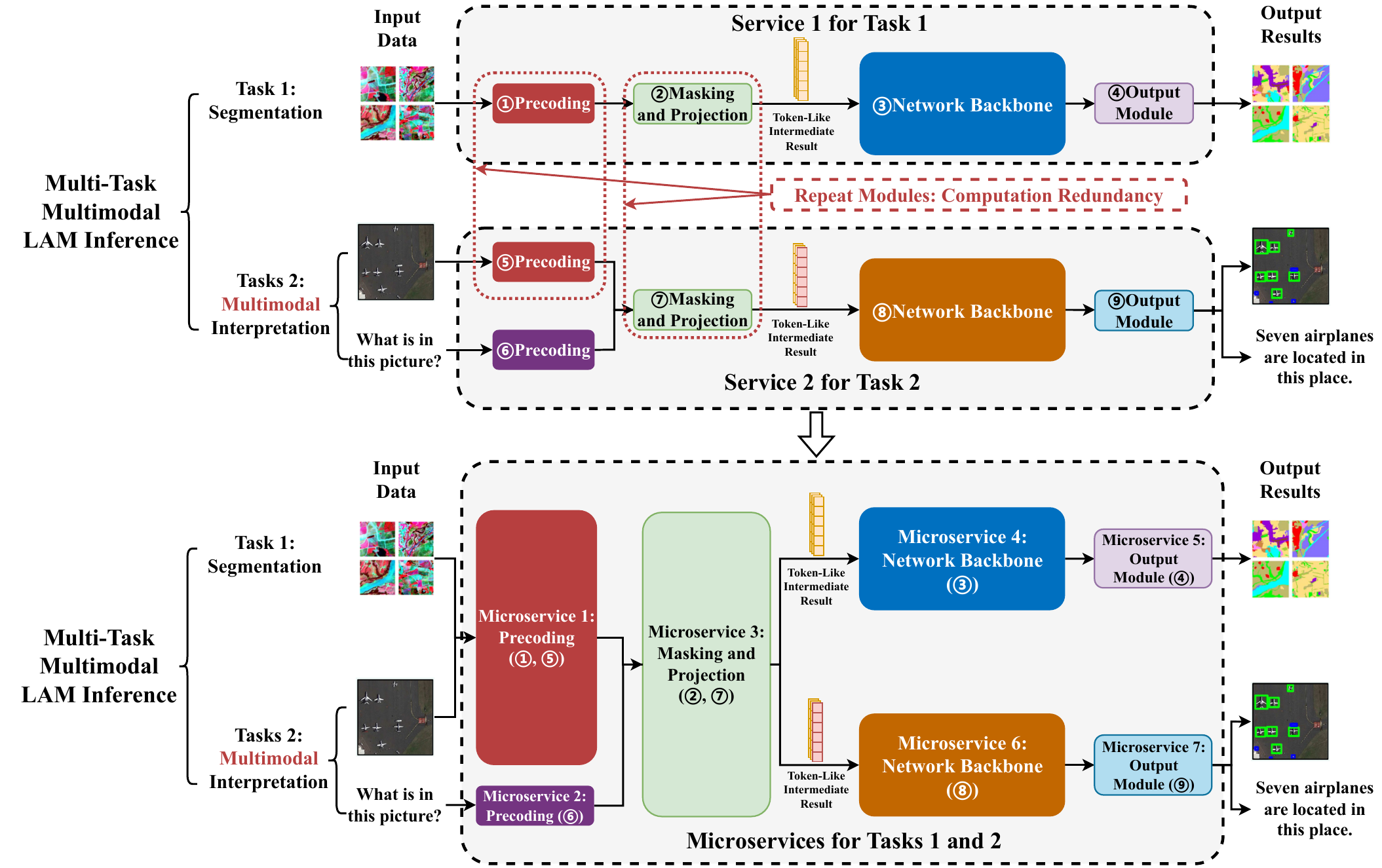}
	\caption[]{Microservice-empowered satellite edge LAM inference architecture. In the presented scenario, precoding and masking and projection modules for tasks 1 and 2 are redundant. As a result, services 1 and 2 can be decomposed into several sequential microservices, with microservices 1 and 3 serving as shared modules for tasks 1 and 2, thereby addressing computation redundancy.}
	\label{fig: microservice}
\end{figure*}

To address the computation redundancy problem and further reduce inference latency, we shall propose a microservice-empowered satellite edge LAM inference architecture for multiple multimodal downstream tasks, which virtualizes the functional modules of the multimodal LAM into a series of independent microservices \cite{yang2024latency}. As illustrated in Fig. \ref{fig: microservice}, traditional service-based multi-task multimodal LAM inference packages the entire computation process into a single service, requiring a complete invocation for each inference task, which clearly overlooks the shared modules between multiple tasks. Specifically, the microservice-empowered satellite edge LAM inference architecture transforms the classical monolithic inference service, typically deployed in data centers, into a unidirectional acyclic graph of distinct microservices, which can be distributed across networks. As shown in Fig. \ref{fig: microservice}, for tasks 1 and 2, microservices 1 and 3 can be simultaneously invoked for the shared modules, followed by utilizing the remaining microservice modules in the original computation sequence to complete the multi-task multimodal satellite edge LAM inference. This architecture offers advantages in terms of agility, portability, scalability, and resilience in software development and maintenance.
Note that frequently packing and unpacking operations over the intermediate activations are performed for exchanges among microservices, the extra communication overhead shall be optimized by a latency aware granularity of microservices scheduling approach while improving the resource utilization.

\begin{figure*}[t]
	\centering
	\includegraphics[width=1\linewidth]{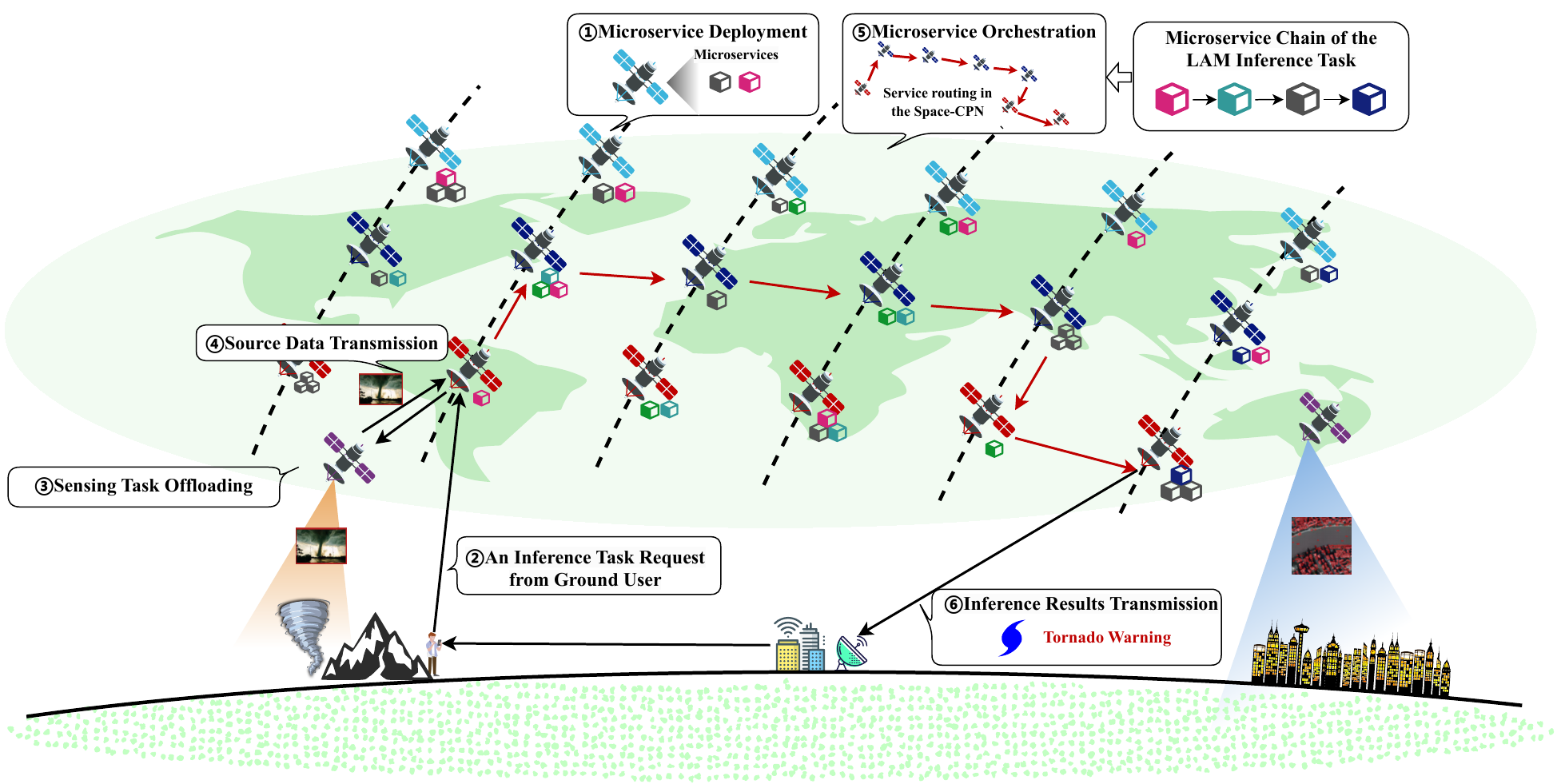}
	\caption[]{Procedures of microservice deployment and orchestration for microservice-empowered satellite edge inference with LAMs including microservice deployment, inference request, sensing task offloading, source data transmission, microservice orchestration, and inference results transmission.}
	\label{fig: satellite}
\end{figure*}

As illustrated in Fig. \ref{fig: satellite}, various microservice modules need to be first pre-deployed on each satellite node.
For each inference request from ground user involving RS data sensing, intelligent inference, and result transmission, the complex computation task is decomposed into multiple submodules, or microservice modules, which are assigned to the corresponding satellite nodes containing the necessary modules for sequential execution. The final results are then transmitted back to the ground user. There are two main challenges in this satellite edge inference process. 
First one is the ``where to compute" issue concerning microservice deployment \cite{yu2024microservice}. Given a known satellite routing table, the goal is to deploy the appropriate microservice modules on each satellite based on the statistical requirements of the inference tasks. Secondly, ``how to compute'' in an issue related to microservice orchestration \cite{guo2024performance}. With the deployed microservice modules, this involves scheduling the computing resources of satellite nodes according to the execution sequence of microservice modules required by real-time inference tasks to complete the computation process.

\subsection{Microservice Deployment and Orchestration}\label{sec: microservice technologies}
Space-CPN provides global coverage, and due to the significant differences in terrain and population density across various regions, the demands for RS and computation services vary greatly. For example, satellites over vast ocean areas may have abundant idle computing resources, while those over densely populated or high-priority monitoring areas face relatively constrained computing resources. Spatially, the task load on satellites varies significantly across different regions. Temporally, each satellite's geographical position changes rapidly, and as it passes over regions with different user densities and RS needs, its task load also fluctuates quickly. Unlike terrestrial networks, idle computing resources in satellite networks are distributed unevenly across both spatial and temporal dimensions. Furthermore, the heterogeneity of satellite computing and communication capabilities (e.g., central processing units (CPUs), memory, energy) in Space-CPN, the uncertainty of the satellite network's dynamic topology, and the interdependencies between different microservice modules make it crucial to efficiently plan and orchestrate computing resources within the network to complete inference tasks effectively.
\subsubsection{Microservice Deployment}\label{sec: dep}
In this article, the microservice deployment in LEO satellite networks refers to the strategic allocation of computation and communication tasks involved in executing multi-task multimodal LAM inference at the satellite edge \cite{yu2024microservice}.
Although performing on-board multi-task multimodal LAM inference offers benefits such as reduced data transmission to Earth and faster response times, the intricate microservice deployment strategies and dynamic satellite topologies raise significant challenges. 
The limited computing power, energy constraints, and unpredictable communication links of LEO satellites further complicate the management of microservices, particularly when handling diverse and heterogeneous inference tasks. 
These inference tasks can be modeled as directed acyclic graphs (DAG), where nodes represent different microservices, and edges denote the data dependencies between them.
{Specifically, the latency in a microservice-empowered inference architecture consists of processing latency, propagation latency, and transmission latency, while the total execution latency for a given task is determined by the slowest path among all microservices involved. 
Suppose multiple inference tasks need to be executed, each of which consists of multiple microservices operating sequentially. The goal of microservice deployment is to optimize the deployment of these microservices within the satellite networks to minimize the total execution latency of all inference tasks while efficiently utilizing the limited resources of heterogeneous satellites.}
The problem can then be modeled as a matching between the DAG of the inference tasks and the time-varying satellite networks for latency minimization.
The resulting optimization problem is a mixed-integer linear programming (MILP) problem due to the integer constraints of deployment indicator, which is a NP-hard problem.

To solve the MILP problem considering the time-varying nature of transmission and propagation latency of ISLs, traditional solvers cannot be directly applied because some objective parameters, such as transmission latency, are functions of optimization variables, which are unknown during the modeling process.
The problem is further reformulated as a Markov decision process (MDP), where the states of the satellite networks, including available resources and communication conditions, are considered.
Moreover, the actions represent deployment choices, and the rewards are based on the latency and resource efficiency of each decision.
We further propose a reinforcement learning (RL)-based solution using proximal policy optimization (PPO). This enables the system to learn how to deploy microservices in a manner that reduces latency and avoids resource bottlenecks.
The use of PPO allows for an adaptable and scalable solution that effectively handles the complex dependencies of microservices across a constellation of LEO satellites. 
The resulting deployment strategies are designed to minimize latency and resource consumption, ensuring efficient execution of multi-task multimodal LAM inference in a resource-constrained and highly dynamic environment. 	{It is worth noting that the microservice deployment problem in this paper can be extended to a multi-agent RL approach \cite{xu2023edge}. Specifically, we can partition the satellite networks into several sub-networks, where each sub-network consists of multiple satellites and is managed by a single agent. The agents communicate with each other to exchange local sub-network information and adjust their strategies accordingly while sharing the same reward function to ensure cooperative decision-making across the satellite networks.	
}

\subsubsection{Microservice Orchestration}\label{sec: orchest}
In this article, we define the microservice orchestration in satellite edge inference architecture as the coordination of computation and communication tasks incurred by designing a service routing scheme for terrestrial requests. The aim of microservice orchestration is to find an energy-efficient microservice routing scheme for these requests over the large-scale mega-constellations with hundreds to thousands of satellites due to the bottleneck on satellites. Specifically, the microservices for satellite edge inference have been deployed on the satellites in advance.
We suppose that there are at least two or more microservices to collaborate together for responding to the request and model the requested service as a DAG to represent the execution processes of microservices. 
We then construct an augmented directed graph, where the weight of each edge represents the energy consumption for both computation and communication when orchestrating microservices over the satellite networks.
{If each satellite in the network is equipped with all possible microservice modules, the microservice orchestration problem will be reduced to a minimum spanning tree problem. However, since it is impractical for each satellite to host all microservice modules, relay nodes must be introduced.
} 
As a result, we can reformulate the problem of energy-efficient microservice routing as a minimum directed Steiner tree (DST) problem to minimize the total energy consumption for satellite edge inference \cite{CHARIKAR1999approximation}. 
We further propose a heuristic algorithm to approximate the optimal solution by first using Dijkstra algorithm to find the minimum distance paths and merging the paths afterwards. 
By solving the DST problem, the resulting microservice orchestration policies can help minimize the total energy consumption of the microservice processing and data transmission procedures for microservice-empowered satellite edge inference.

While heuristic and approximating algorithms for solving the DST problem are suitable for static topologies, they may perform poorly when applied to highly dynamic networks with time-varying topologies like satellite networks. It is thus essential to explore solutions for the DST problem under time-varying topologies.
Traditional neural networks struggle to capture the structural information of nodes and edges in a graph, whereas graph neural networks (GNNs), a type of neural network specifically designed to process graph data, can effectively extract useful structural features through the message-passing mechanism between neighboring nodes. 
In addition, RL can incrementally build partial solutions by learning a policy to make optimized choices during the decision-making process without requiring pre-labeled data. 
Consequently, combining RL with GNNs becomes an ideal choice for solving the DST problem in the presence of time-varying topologies \cite{benjamin2022deep}. In the DST problem, RL is modeled as a MDP, with the following elements. State space is the state of the current partial solution on the graph, including selected nodes and edges, graph information, and problem information. 
Action space is the selection of a new node to add to the current solution in the partial solution. 
Reward function is the reward at each step equals the negative weight of the selected edge.
Policy selects actions based on the estimated Q-value function, where the Q-value represents the expected future cumulative reward after taking a particular action. For the choice of deep Q-Network, the graph attention network (GAT) can be employed because it can capture the relationships between nodes in a graph and update each node's embedding through message-passing mechanisms. 
In the solving process, we input the graph-structured data, i.e., the state, into the GAT, and the output is the action for network decision-making.

\section{Future Directions}\label{sec: future}
In this section, we discuss future directions for satellite edge LAM, encompassing the following three aspects: task-oriented communications for downstream tasks, more energy-efficient computing architectures, and the application of generative AI in optimizing satellite edge AI networks.

\subsection{Task-Oriented Communications for Satellite Edge LAM}
Task-oriented communication provides a solution for satellite edge AI inference by extracting key features and eliminating redundant information, thereby improving the efficiency of information transmission between satellites and GSs \cite{shi2023task}. 
In terms of satellite edge AI, satellites are equipped with various sensors that capture multimodal data, such as optical imagery, infrared, and radar signals. Unlike traditional multimodal data, RS multimodal data exhibits distinct characteristics, often being multi-spectral and capturing observations from various angles. 
The resolution of data collected by different sensors can vary significantly. 
Satellites can also observe the same location over extended periods, providing a valuable temporal dimension. 
These multimodal RS data are capable of providing high-quality features for downstream tasks such as change detection and cloud removal. However, existing data interpretation methods suffer from several limitations. 
First, traditional methods require downloading raw data to GSs for post-processing, which is at odds with the limited spectral resources of SGLs. 
Second, traditional single-sensor, data-driven, unimodal deep learning methods struggle to achieve optimal performance when applied to different modalities. 
Furthermore, traditional multimodal approaches often fail to effectively align data based on each modality's characteristics, making it difficult to fuse data from different imaging mechanisms and extract meaningful insights. Finally, meteorological effects, such as atmospheric turbulence, pose significant challenges for the data sensing, processing procedures and the transmission in SGLs \cite{Bui2024semantic}.
To address these challenges, we propose a solution based on the multimodal information bottleneck (MIB) for RS downstream tasks that aims to reduce the volume of data transmitted through SGLs while maximizing the extraction of features from multimodal RS data. {The information bottleneck \cite{tishby1999information} is originally designed as a task-oriented communication method that effectively characterizes the trade-off between informativeness and compression. It aims to extract features that are highly informative for downstream tasks while minimizing task-irrelevant information, thereby reducing transmission overhead.}

The MIB approach provides a robust framework for this purpose, enabling dimensionality reduction within each feature space to minimize information loss, followed by aligning heterogeneous output features for more effective fusion \cite{mai2022multimodal}. MIB involves first finding a compressed representation for each input modality that maximizes the retention of information relevant to the target task while minimizing redundancy. These compressed representations are then fused into an efficient and robust encoded feature vector. The GSs receive this noise-corrupted feature vector to derive the inference result for downstream tasks.
MIB focuses on extracting the most relevant features from each modality that are crucial for a specific task. 
By emphasizing task-relevant features, the model effectively reduces the dimensionality of feature vectors, which is particularly advantageous in bandwidth-constrained satellite communication environments. 
In SGLs, noise can significantly distort the transmitted signal. 
We introduce the variational information bottleneck approach for maintaining high-quality inference capabilities even when the transmitted feature vectors are significantly corrupted by noisy environments.
This approach improves the reliability of data transmission and stabilizes communication, even in the face of environmental disturbances \cite{xie2023robust}.

\subsection{Neuromorphic Computing for Energy-Efficient Satellite Edge LAM}
Energy consumption is becoming the main bottleneck for LAM-empowered computation tasks, and the situation is even worse for satellites, who have limited energy-harvesting capabilities and constrained batteries lifespan \cite{yang2016towards}. As a result, a more energy-efficient neural network model for LAMs is essential.
To this end, spiking neural network (SNN), the third-generation neural network, has exhibited a great potential for LAMs with its unique spike-based event-driven mechanism \cite{schuman2022opportunities}. 
Unlike traditional neural networks, it generally uses leaky integrate-and-fire model to characterize neurons, where a gradually decreased membrane potential is adopted to record the temporal dynamics. 
The weighted sum of the previous layer's output contributes to such membrane potential during forward propagation, and each neuron shall emit a spike as output if it's membrane potential reaches a threshold. 
Based on this special neuron structure, the computation flow can be described as the transition of sparse, discrete, and single-bit spikes.
By mimicking human neuron's behavior with extreme sparse computation, the architecture of LAMs can be re-designed to a more energy-efficient version. 
Specifically, in \cite{yao2023spike}, a spike-driven transformer architecture is proposed for the basic block of the large language models. 
As for vision generation tasks where diffusion models are widely employed, SNN is also attractive due to its biological plausibility compared to traditional artificial neural network \cite{cao2024spiking}. 
Overall, these existing architectures convince the fact that SNN is a promising solution for energy-efficient LAMs.

Another way for addressing the energy consumption issue is to developing innovative hardware computing architectures.
Traditionally, AI models are mainly deployed and executed by advanced modern microprocessors such CPUs and GPUs, which offer a strong computation capability for complex AI tasks.
To fully unleash the potential of these aforementioned SNN-based architectures in energy efficiency, neuromorphic processors is a necessary part acting as the hardware for model deployment \cite{schuman2022opportunities}. 
Different from classical von Neumann architecture where processing and storage capabilities are separately enabled by CPUs and memory units, the computation task is completed via membrane potential dynamics on neurons and asynchronous binary spikes on synapses, which effectively avoids the throughput limitation caused by von Neumann bottleneck.
{In summary, brain-inspired SNNs significantly reduce computational complexity and enhance energy efficiency due to their binarized input and output. Additionally, their event-driven nature enables the explicit encoding of input and output into time-dimensioned spike sequences for processing, effectively extracting temporal features. This makes neuromorphic computing well-suited for satellite edge AI for tasks like on-board environmental monitoring \cite{bose2016spiking}.}
The Neuro SatCom project that put forward by the European Space Agency, have already evaluate the employment of neuromorphic computing to on-board applications. Specifically, for an interference detection task, AI models on the non-neuromorphic platform Xilink Versal consume 136.9 J per problem while on the neuromorphic platform Spinnaker only consume 6.3 J per problem. Therefore, neuromorphic computing is particularly suitable for resource-limited satellites, which can be further designed to support LAMs \cite{lagunas2024performance}.

\subsection{Generative AI for Optimizing Satellite Edge AI Networks}
The allocation of communication, computation, storage, and sensing resources plays a critical role in the efficiency of satellite edge AI networks \cite{sheng2023coverage}. 
To enhance the efficiency of satellite edge AI systems, finding effective solutions for large-scale resource optimization problems is of paramount importance.
The development of generative AI offers new approaches to solving optimization problems in satellite edge AI systems. For instance, the microservice deployment problem discussed in Section \ref{sec: dep}, which can be formulated as a MILP problem, can be solved using a diffusion model \cite{sanokowski2024diffusion}. The process begins by formulating the original problem as an energy minimization task. During the forward diffusion phase, noise is added to an initial near-valid solution, transforming it into a more randomized form, allowing exploration of the solution space. The reverse diffusion phase then removes noise in small steps to reconstruct a solution close to the optimum, driven by minimizing reverse Kullback-Leibler divergence. GNNs are also integrated to capture complex relationships in graph-structured data, improving solution quality. Annealing techniques are used to gradually reduce temperature, reducing randomness and focusing on low-energy regions for convergence toward a global optimum.
As another example, solving the DST problem under time-varying topologies, as discussed in Section \ref{sec: orchest}, can also be approached using generative AI methods. Specifically, by leveraging the Transformer's ability to capture the relevance of historical information, the model can better utilize the correlation between satellite topologies at adjacent time steps, selecting actions to optimize the reward in the RL-based solution to the DST problem \cite{hu2024on}.
The core challenge of this approach is how to encode graph-structured data into semantically meaningful vector representations for input into the Transformer. To address this, the encoding principles of Graph Transformer can be used to encode graph-structured data \cite{ying2021transformers}, thereby replacing the GAT-based Deep Q-Network with a Transformer architecture.

Although these methods have achieved certain success in solving their respective optimization problems in satellite edge AI systems, they are limited to addressing the current problem, i.e., the specific data and environment distribution, and cannot be generalized to solve other problems. 
Therefore, we attempt to explore a ``one model for all problems'' learning model, utilizing LAMs to provide a unified solution framework for large-scale satellite AI system optimization problems.
The authors in \cite{cui2024netllm} proposed NetLLM, an large language model (LLM) adaptation framework for networking, which is a specific category of LAMs. It integrates a multimodal encoder for networking data and introduces networking-specific heads for task-specific outputs. The data-driven low-rank networking adaptation reduces fine-tuning costs by adding lightweight, low-rank matrices without altering the pre-trained LLM. NetLLM effectively handles tasks such as viewport prediction, adaptive bitrate streaming, and cluster job scheduling.
To this end, LAMs specifically designed for satellite AI system optimization appear to be highly promising for microservice deployment, orchestration, and migration.
Moreover, LLMs can also be considered as solvers for large-scale optimization problems. 
LLMs present a promising avenue for democratizing optimization by allowing individuals to describe optimization tasks in natural language, reducing the need for specialized expertise. However, LLMs face challenges in mathematical optimization, including limited context windows that restrict the detailed representation of large-scale problems with multiple constraints, and difficulty in navigating non-convex or highly irregular search spaces \cite{yang2024large}. 
To address these limitations, recent studies in the academic community have begun exploring various strategies. In \cite{ahmaditeshnizi2024optimus}, the authors proposed a LLM-based agent framework called OptiMUS to simplify the use of domain-specific tools, such as Gurobi, through text prompts. Similarly, the authors in \cite{zhang2024solving} proposed an OptLLM framework that employs multi-round dialogues to iteratively refine the modeling and solving of optimization problems.
Unlike the aforementioned works that trained a LLM for one or a few classes 
of MILP problems, the authors in \cite{li2024towards} proposed to train a single foundation model on diverse MILP classes and instances, thereby improving the generalization ability to unseen classes.
In summary, fully exploring and harnessing the potential of LLMs as solvers for satellite AI system optimization problems is an another significant direction for future research.

\section{Conclusions}
Satellite edge AI represents a new paradigm for space data processing, shifting the focus from ground-based to on-board processing techniques. 
This paradigm involves disruptive technologies for fine-tuning and inference of LAMs at the satellite edge.
To enhance the timeliness, effectiveness, and trustworthiness of RS downstream tasks for LAM-empowered satellite edge AI, developing computation task decomposition principles is the key to support scalable LAM deployment in resource-constrained Space-CPN with dynamic topology.
This article contributed to propose a satellite federated fine-tuning architecture for fine-tuning LAMs by partitioning and distributing LAM modules and developing corresponding inter-satellite link scheduling and space-ground coordinated transmission schemes. 
Additionally, we introduced a microservice-empowered satellite edge LAM inference architecture to address the computational redundancy problem in multi-task multimodal LAM inference and provided insightful solutions for microservice deployment and orchestration problems at the satellite edge.
We outlined our future works from three perspectives: task-oriented communications, neuromorphic computing, and generative AI for optimizing satellite edge AI networks. 
We hope this article will inspire more innovative architectures, technologies, and applications of LAMs for satellite edge AI.
\bibliographystyle{IEEEtran}
\bibliography{reference}


\end{document}